%% file: sample-sigconf.tex
\documentclass[sigconf]{acmart}

\usepackage{booktabs} % For formal tables
\usepackage{adjustbox}

\setcopyright{none}
\graphicspath{{./figures/}}

\begin{document}
\title{Recurrent Neural Networks for Person Re-identification Revisited}

%\author{Anonymous submission\\Paper ID 128}

\author{Jean-Baptiste Boin}
\affiliation{
  \institution{Stanford University}
  \city{Stanford}
  \state{CA}
  \country{U.S.A.}
}
\email{jbboin@stanford.edu}

\author{Andr\'e Araujo}
\affiliation{
  \institution{Google AI}
  \city{Mountain View}
  \state{CA}
  \country{U.S.A.}
}
\email{andrearaujo@google.com}

\author{Bernd Girod}
\affiliation{
  \institution{Stanford University}
  \city{Stanford}
  \state{CA}
  \country{U.S.A.}
}
\email{bgirod@stanford.edu}

\begin{abstract}
The task of person re-identification has recently received rising attention due to the high performance achieved by new methods based on deep learning. In particular, in the context of video-based re-identification, many state-of-the-art works have explored the use of Recurrent Neural Networks (RNNs) to process input sequences. In this work, we revisit this tool by deriving an approximation which reveals the small effect of recurrent connections, leading to a much simpler feed-forward architecture. Using the same parameters as the recurrent version, our proposed feed-forward architecture obtains very similar accuracy. More importantly, our model can be combined with a new training process to significantly improve re-identification performance. Our experiments demonstrate that the proposed models converge substantially faster than recurrent ones, with accuracy improvements by up to 5\% on two datasets. The performance achieved is better or on par with other RNN-based person re-identification techniques.
\end{abstract}

%\begin{CCSXML}
%<ccs2012>
%  <concept_id>10010520.10010553.10010562</concept_id>
%  <concept_desc>Computer systems organization~Embedded systems</concept_desc>
%  <concept_significance>500</concept_significance>
%  <concept_id>10010520.10010575.10010755</concept_id>
%  <concept_desc>Computer systems organization~Redundancy</concept_desc>
%  <concept_significance>300</concept_significance>
%  <concept_id>10010520.10010553.10010554</concept_id>
%  <concept_desc>Computer systems organization~Robotics</concept_desc>
%  <concept_significance>100</concept_significance>
%  <concept_id>10003033.10003083.10003095</concept_id>
%  <concept_desc>Networks~Network reliability</concept_desc>
%  <concept_significance>100</concept_significance>
%</ccs2012>
%\end{CCSXML}

%\ccsdesc[500]{Computer systems organization~Embedded systems}


\maketitle

\input{samplebody-conf}

\bibliographystyle{ACM-Reference-Format}
\bibliography{sample-bibliography}

\end{document}

%% file: samplebody-conf.tex
\section{Introduction}

Person re-identification consists of associating different tracks of a person as they are captured across a scene by different cameras. There are many applications for this task. The most obvious one is video-surveillance. It is common in public spaces to deploy networks of cameras with non-overlapping field of views that capture different areas. These networks produce large amount of data and it can be very time-consuming to manually analyze the video feeds to keep track of the actions of a single person as they move across the various fields of view. Person re-identification allows this task to be automated and makes it scalable to keep track of the trajectories of a high number of different identities. Solving this problem can also be critical for home automation, where it is important to keep track of the location of a user as they move across the different rooms, for single-camera person tracking in order to recover from occlusions, or for crowd dynamics understanding, among other tasks. The challenges inherent to this task are the variations in background, body pose, illumination and viewpoint. It is important to represent a person using a descriptor that is as robust as possible to these variations, while still being discriminative enough to be characteristic of a single person's identity.

A sub-class of this problem is \emph{video-based} re-identification, where the goal is to match a video of a person against a gallery of videos captured by different cameras, by opposition to image-based (or single-shot) re-identification, where only a single view of a person is provided.

Person re-identification has recently received rising attention due to the much improved performance achieved by methods based on deep learning. For video-based re-identification, it has been shown that representing videos by aggregating visual information across the temporal dimension was particularly effective. Recurrent Neural Networks (RNNs) have shown promising results for performing this aggregation in multiple independent works \cite{McLaughlin16, yan2016person, wu2016deep, zhou2017see, zhang2017learning, xu2017jointly, chen2017deep}. In this paper, we analyze one type of architecture that uses RNNs for video representation. The contributions of this work are the following. We show that the recurrent network architecture can be replaced with a simpler non-recurrent architecture, without sacrificing the performance. Not only does this lower the complexity of the forward pass through the network, making the feature extraction easier to parallelize, but we also show that this model can be trained with an improved process that boosts the final performance while converging substantially faster. Finally, we obtain results that are on par or better than other published work based on RNN, but with a much simpler technique.

\section{Related work}

The majority of the traditional approaches to image-based person re-identification follow a two-step strategy. The first step is feature representation, which aims at representing an input in a way that is as robust as possible to variations in illumination, pose and viewpoint. Here is a non-exhaustive list of some of the more prominent techniques commonly used to craft such representations: Scale Invariant Feature Transforms (SIFT) \cite{lowe2004distinctive} used in \cite{zhao2013person}, \cite{zhao2013unsupervised}, Scale Invariant Local Ternary Patterns (SILTP) \cite{liao2010modeling} used in \cite{liao2015person}, Local Binary Patterns \cite{ojala2002multiresolution} used in \cite{xiong2014person}, \cite{yan2016person}, color histograms used in \cite{liao2015person}, \cite{xiong2014person}, \cite{zhao2013person}, \cite{zhao2013unsupervised}, \cite{yan2016person}. This step is followed by metric learning. Using training data, the features are transformed in a way that maximizes intra-class similarities while minimizing the inter-class similarities. Some examples of metric learning algorithms that were specifically introduced for person re-identification are Cross-view Quadratic Discriminant Analysis (XQDA) \cite{liao2015person}, Local Fisher Discriminant Analysis (LFDA) \cite{pedagadi2013local}, based on Fisher Discriminant Analysis (FDA) \cite{fisher1936use}, and its kernelized version k-LFDA \cite{xiong2014person}. See \cite{zheng2016person} for a more detailed survey of these techniques.

More recently, with the successes of deep learning in computer vision \cite{krizhevsky2012imagenet}, \cite{simonyan2014very}, as well as the release of larger datasets for re-identification (VIPeR \cite{gray2008viewpoint},  CUHK03 \cite{li2014deepreid}, Market-1501 \cite{zheng2015scalable}), this field has shifted more and more towards neural networks. 

In particular, the Siamese network architecture \cite{bromley1994signature}, \cite{hadsell2006dimensionality} provides a straightforward way to simultaneously tackle the tasks of feature extraction and metric learning into a unified end-to-end system. This architecture was introduced to the field of person re-identification by the pioneering works of \cite{yi2014deep} and \cite{li2014deepreid}. This powerful tool can learn an embedding where inputs corresponding to the same class (or identity) are closer to each other than inputs corresponding to different classes. It also has the added benefit that it can be used even if a low number of images is available per class (such as a single pair of images), unlike classification approaches that would require more data. Different variants of the Siamese network have been used for re-identification. \cite{ahmed2015improved} achieved very good results by complementing the Siamese architecture with a layer that computes neighborhood differences across the inputs. Instead of using pairs of images as inputs, \cite{cheng2016person} uses triplets of images whose representations are optimized by using the triplet loss that was first used in \cite{schroff2015facenet} for embedding tasks.

Although slightly less explored, the topic of video-based re-identification has followed a similar path since many techniques from image-based re-identification are applicable, ranging from low-level hand-crafted features \cite{liu2015spatio}, \cite{wang2014person} to deep learning, made possible by the release of large datasets (PRID2011 \cite{hirzer2011person}, iLIDS-VID \cite{wang2014person}, MARS \cite{zheng2016mars}). In order to represent a video sequence, most works consider some form of pooling that aggregates frame features into a single vector representing the video. Some approaches such as \cite{zheng2016mars} do not explicitly make use of the temporal information, but other works have shown promising results when learning spatio-temporal features. In particular, \cite{McLaughlin16}, \cite{yan2016person}, \cite{wu2016deep} all propose to use Recurrent Neural Networks (RNNs) to aggregate the temporal information across the duration of the video. Even though these works use different recurrent architectures (resp. vanilla RNN, LSTM (Long short-term memory) and GRU (Gated Recurrent Unit)), they are related to each other. \cite{zhou2017see} showed promising results by combining a RNN-based temporal attention model with a spatial attention model.

In this work, we will focus more in detail on \cite{McLaughlin16}, which directly inspired more recent papers that built upon it: \cite{zhang2017learning} replaces the RNN with a bi-directional RNN; \cite{xu2017jointly} computes the frame-level features with an extra spatial pyramid pooling layer to generate a multi-scale spatial representation, and aggregates these features with a RNN and a more complex attentive temporal pooling algorithm; \cite{chen2017deep} aggregates the features at the output of the RNN with the frame-level features used as the input to the RNN, and also processes the upper-body, lower-body and full-body sequences separately, with late fusion of the three sequence descriptors. All propose a more complex system compared to \cite{McLaughlin16}, but showed improved performance.

\section{Proposed framework}

\subsection{General architecture of the network}

In video-based person re-identification, the goal is to associate videos of the same person, which may be taken with different cameras. A query video of a person can be matched against a gallery of videos, either by using a multi-match strategy where frame descriptors from query and gallery video are compared pairwise, or by a single-match strategy where the information is first pooled across all frames to represent each video with a single high-dimensional descriptor. The latter strategies have slowly superseded the former ones and have proved more successful, on top of being more efficient.

\begin{figure}
\includegraphics[width=0.8\linewidth]{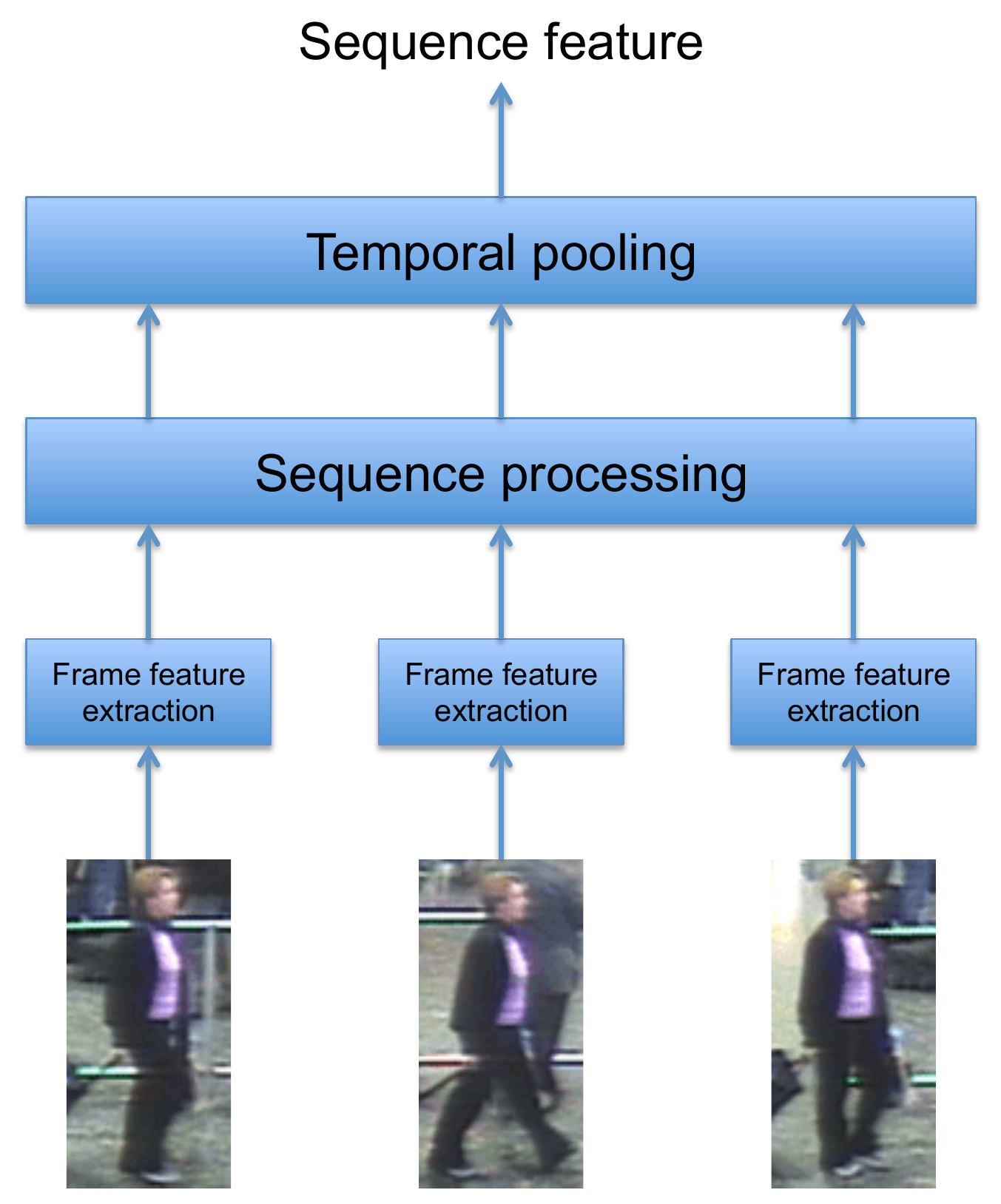}
%\vspace{-6mm}
\caption{General architecture of the feature extraction network.}
\label{fig_architecture}
\end{figure}

In order to extract a fixed length one-dimensional descriptor from a variable-length sequence of images, we introduce a parametric model called the feature extraction network. The general architecture of that network is shown in Fig. \ref{fig_architecture}: it is made up of three distinct stages namely frame feature extraction, sequence processing and temporal pooling. This multiple-stage architecture is considered because is a good generalization of the systems used in the related works \cite{McLaughlin16} (as well as its extensions \cite{zhang2017learning, xu2017jointly, chen2017deep}), \cite{yan2016person} and \cite{wu2016deep} for video representation.

The first stage (frame feature extraction) independently extracts a descriptor of dimension $d_1$ for each frame of the input video. This descriptor should capture the appearance information that needs to be aggregated across the sequence.

The second stage (sequence processing) takes the sequence of frame descriptors and outputs a (different) sequence of fixed-length vectors of dimension $d_2$ (it is possible to choose $d_2 \neq d_1$). In general, this stage mixes the information contained in all the frames of the sequence. One example would be to compute the pairwise differences between consecutive frame descriptors, but this stage can perform more complex operations.

Last, the temporal pooling stage outputs a single fixed-length vector from the variable-length sequence at the output of the previous stage. Note that in the general formulation, this temporal pooling could depend on the order of the input sequence, but in general much simpler strategies are used: \cite{McLaughlin16} only considers max and average pooling, showing that average pooling generally outperforms max pooling, and \cite{yan2016person} and \cite{wu2016deep} both use average pooling only. Other works explored more complex temporal pooling strategies \cite{xu2017jointly, zhou2017see}, but in this work only average pooling, which happens to be order-independent, is considered.

Here we will only focus on the specific architecture used in \cite{McLaughlin16}, which has shown great success. The frame feature extraction stage is a convolutional neural network (CNN), the sequence processing stage is a recurrent neural network (RNN) and the temporal pooling is performed by average pooling the outputs at each time step. The full network is trained using the Siamese architecture framework.

We call $f^{(t)} \in \mathbf{R}^{d_1}$ ($o^{(t)} \in \mathbf{R}^{d_2}$) the inputs (outputs) of the sequence processing stage, $t = 1, ..., T$. The output at each time step is given by the RNN equations:
\begin{align}
\label{eq_rnnoutput}
o^{(t)} &= W_i f^{(t)} + W_s r^{(t-1)}
\end{align}
where $r^{(t-1)} = Tanh\left(o^{(t-1)}\right)$ is obtained from the previous output.

The last output of the RNN technically contains information from all the time steps of the input sequence so it could be used to represent that sequence (the temporal pooling stage would then just ignore $o^{(t)}$ for $t \leq T-1$ and directly output $o^{(T)}$). However, in practice if we examine the contribution of a given input to each of the time steps of an output of a RNN, usually this contribution decreases over time, as the information of the earlier time steps gets diluted and the later time steps dominate. This means that the value of $o^{(T)}$ is much more strongly dependent on $i^{(T)}$ than on $i^{(1)}$. This is not suitable when designing a sequence descriptor that should be as representative of the start of the sequence as of its end. In order to prevent this phenomenon, \cite{McLaughlin16} shows that a good choice is to use average pooling as a temporal pooling stage. A sequence can thus be represented as:
\begin{align}
\label{eq_meanpool}
v_s = \frac{1}{T} \sum_{t=1}^T o^{(t)}
\end{align}

\subsection{Proposed feed-forward approximation} \label{sec_simplification}

This limitation of RNNs regarding long-term dependencies has been widely studied and is related to the problem of vanishing gradients. Motivated by this phenomenon, we propose to approximate the RNN to consider only a one step dependency. Under this approximation, the effect of inputs $f^{(1)}, ..., f^{(t-2)}$ on $o^{(t)}$ is in general negligible for a trained RNN. In other words, $o^{(t)}$ may only depend on the previous and the current time steps: $f^{(t-1)}$ and $f^{(t)}$. Hence, \eqref{eq_rnnoutput} can then be rewritten as:
\begin{align*}
o^{(t)} &= W_i f^{(t)} + W_s Tanh\left(o^{(t-1)}\right) \\
&= W_i f^{(t)} + W_s Tanh\left(W_i f^{(t-1)} + W_s r^{(t-2)}\right) \\
&\approx W_i f^{(t)} + W_s Tanh\left(W_i f^{(t-1)}\right)
\end{align*}

The appearance descriptor of the sequence from \eqref{eq_meanpool} can now be approximated as
\begin{align*}
v_s &= \frac{1}{T} \sum_{t=1}^T o^{(t)} \\
&\approx \frac{1}{T} \sum_{t=1}^T  \left( W_i f^{(t)} + W_s Tanh\left(W_i f^{(t-1)} \right) \right) \\
&= \frac{1}{T} \sum_{t=1}^T  W_i f^{(t)} + \frac{1}{T} \sum_{t=0}^{T-1}  W_s Tanh\left(W_i f^{(t)} \right) \\
&= \frac{1}{T} \sum_{t=1}^T  \left( W_i f^{(t)} + W_s Tanh\left(W_i f^{(t)} \right) \right) - \frac{1}{T} W_s Tanh\left(W_i f^{(T)} \right) \\
\end{align*}

For large enough values of $T$, the contribution of the last term is small compared to the sum, so it can be neglected. We now introduce $\tilde{o}^{(t)} = W_i f^{(t)} + W_s Tanh\left(W_i f^{(t)} \right)$, which only depends on $f^{(t)}$. Under the assumptions that were made, we can rewrite
\begin{align}
\label{eq_meanpoolapprox}
v_s &\approx \frac{1}{T} \sum_{t=1}^T \tilde{o}^{(t)}
\end{align}

\begin{figure}
\includegraphics[width=0.75\linewidth]{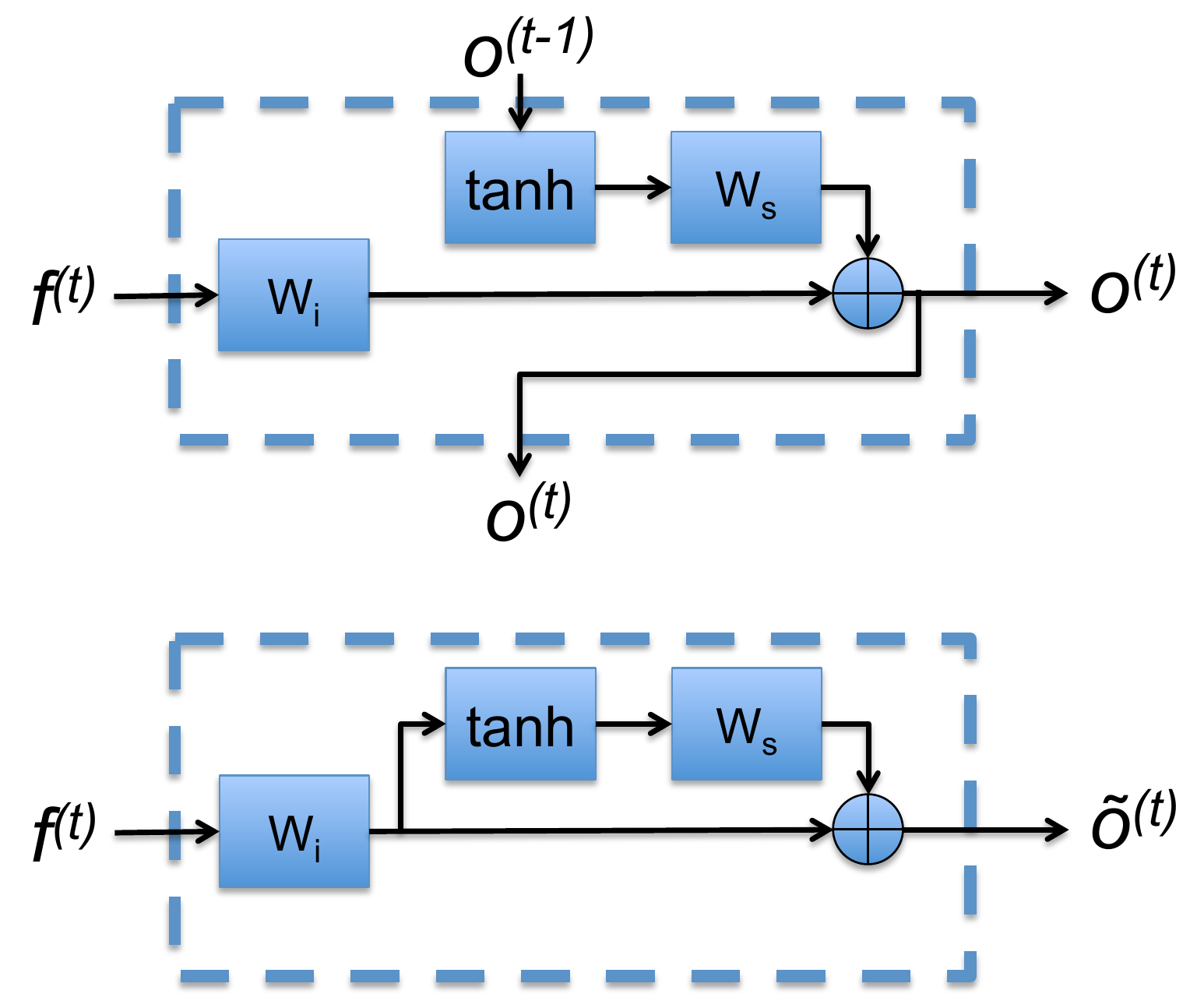}
%\vspace{-6mm}
\caption{Architecture of the sequence processing stage. Top: Recurrent architecture used as baseline (RNN). Bottom: Our non-recurrent architecture (FNN). The blocks labeled $W_i$ and $W_s$ correspond to fully connected layers.}
\label{fig_seqprocarchitecture}
\end{figure}

In other words, the sequence processing stage can be approximated by a simpler non-recurrent network if the RNN is swapped with the feed-forward network shown in Fig. \ref{fig_seqprocarchitecture}. From now on, this architecture of the sequence processing stage will be referred to as FNN (feed-forward neural network). It is important to note that both architectures have the same number of parameters, so their memory footprint is the same.

In order to keep the equations simple the bias terms were ommitted to improve clarity, but a very similar approximation can be derived when they are taken into consideration using the full equation:
\begin{align*}
o^{(t)} &= W_i f^{(t)} + b_i + W_s Tanh\left(o^{(t-1)}\right) + b_s
\end{align*}

In this case, we still ignore the recurrent part when substituting $o^{(t-1)}$ with its expression
\begin{align*}
o^{(t)} &\approx W_i f^{(t)} + b_i + W_s Tanh\left(W_i f^{(t-1)} + b_i\right) + b_s
\end{align*}
and we then perform the second approximation which again gives us the equation  \eqref{eq_meanpoolapprox} with $\tilde{o}^{(t)} = W_i f^{(t)} + b_i + W_s Tanh\left(W_i f^{(t)} + b_i \right) + b_s$.

It is interesting to notice that our proposed FNN architecture contains a shortcut connection, reminiscent of a ResNet block, as was introduced in \cite{he2016deep}. In fact, preliminary experiments showed that this residual connection is critical to the success of the training of this network and removing it causes a considerable drop in the re-identification performance.

\subsection{Improved training pipeline} \label{sec_trainingmodes}

Compared to the RNN, our proposed FNN architecture keeps the same inputs and outputs, so it can be trained just like the RNN by using a Siamese network. The loss that is used is unchanged (combination of a contrastive loss for a pair of inputs and of an identification loss for each input of the pair).

It is impossible to train an RNN with input sequences of length 1. However, by removing the time dependency, our reformulation of the sequence processing stage as a FNN removes this constraint and allows us to process individual frames as sequences of length 1. Computing the representation of a sequence of length $L$ requires about the same amount of computation time and memory as computing the representation of $L$ individual frames.

\begin{figure}
\includegraphics[width=\linewidth]{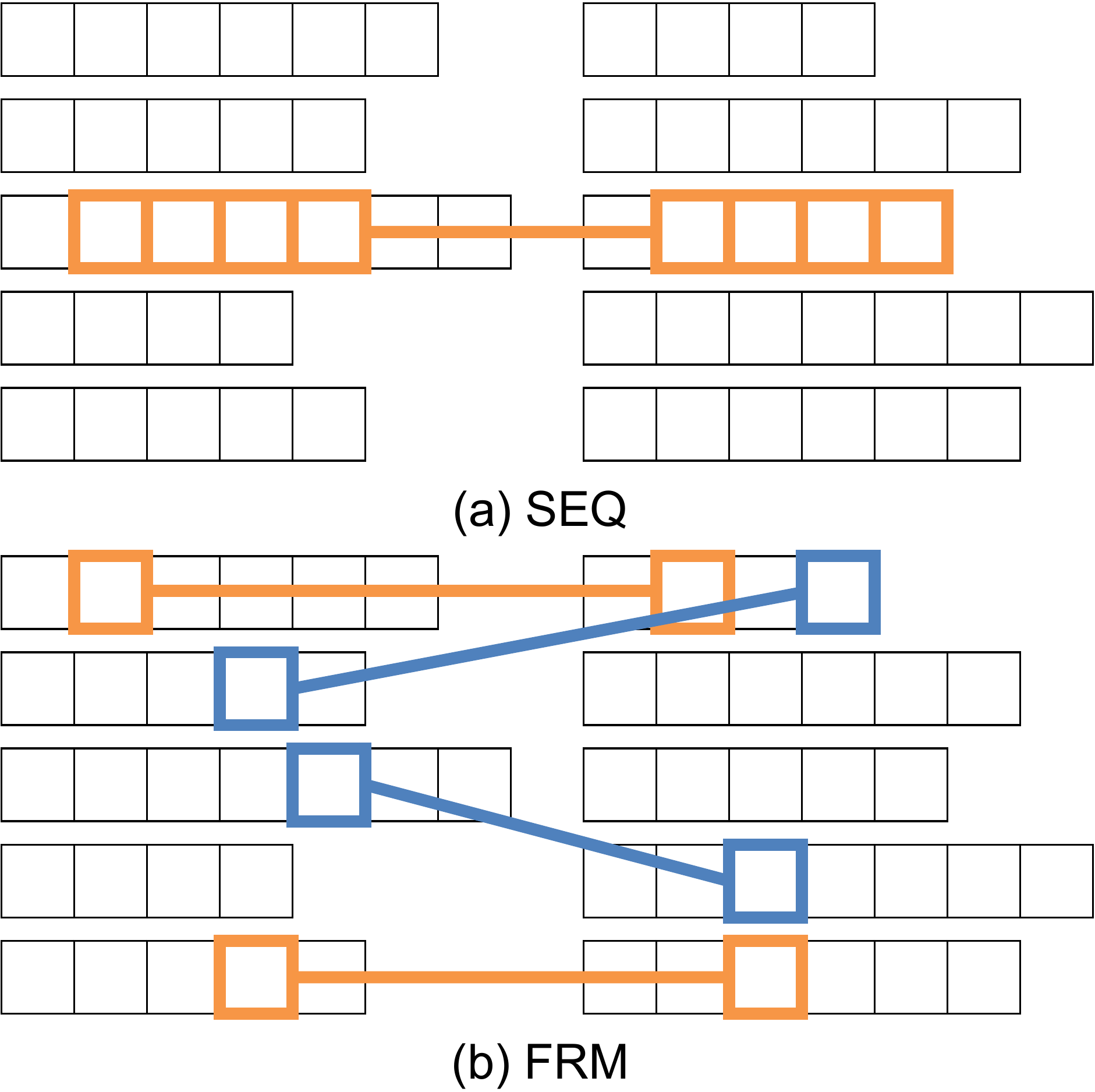}
\includegraphics[width=\linewidth]{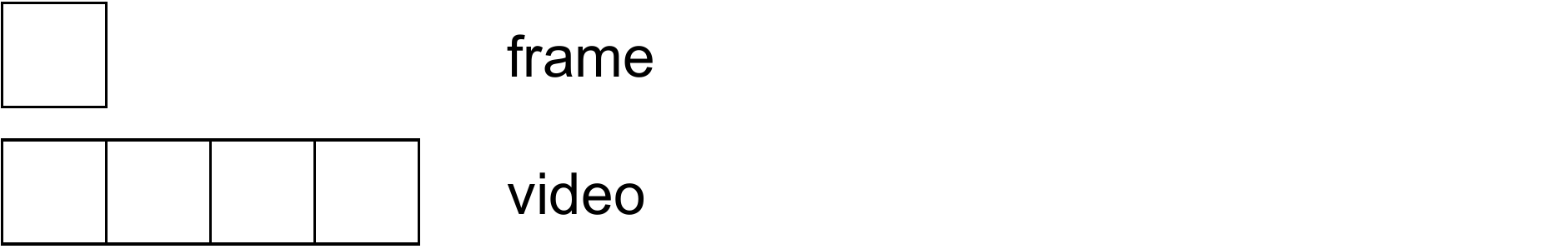}
\caption{Illustration of the sampling strategy for the two training modes: SEQ (sequence as an input) and FRM (individual frames as an input). The dataset is represented in black: each row represents the two videos available for a given person. We consider a mini-batch with the settings $B = 1$ and $L = 4$. The frames loaded in the mini-batch are the ones with the colored outline. The linked sequences or frames correspond to the pairs used as inputs to the Siamese network. Positive pairs (same identity) are shown in orange; negative pairs (different identity) in blue. SEQ only samples adjacent frames from a maximum of $2B = 2$ distinct sequences while FRM can sample unconstrained frames from up to $2BL = 8$ sequences.}
\label{fig_trainmodes}
\end{figure}

We denote $B$ the size of our mini-batch when training the RNN and $L$ the length of the sequences used for training. When training in sequence mode (noted SEQ), within a mini-batch, it is required to load $2 B L$ images (the factor $2$ is due to the Siamese architecture) from up to $2 B$ distinct sequences. If individual frames are used as inputs for training instead (frame mode, noted FRM), it is possible to load and process $B L$ pairs of images with roughly the same memory requirement. This means that a FRM mini-batch is much more diverse since it can now include images from $2 B L$ distinct sequences. The two modes are illustrated in Fig. \ref{fig_trainmodes}.

\section{Experiments}

\subsection{Data and experimental protocol}

\indent \indent \textbf{Datasets.}
We compare the network architectures and training modes on two datasets: the PRID2011 \cite{hirzer2011person} and the iLIDS-VID \cite{wang2014person} datasets. Both datasets were created from videos of pedestrians observed in two non-overlapping camera views, and the matching identities on both views are provided.

Two video sequences of varying length are available for each of these identities. There are 200 such identities (pairs of video sequences) in the PRID2011 dataset, and 300 in the iLIDS-VID dataset. Note that the PRID2011 dataset contains additional distractor videos of people appearing in only one of the view. We follow the standard testing protocol for this dataset as used in \cite{wang2014person}: distractor videos are ignored and only the 200 pairs of videos corresponding to matching identities are considered. Both datasets have comparable video sequence lengths: they range from 5 to 675 with an average of 100 frames for the PRID2011 dataset, and from 22 to 192 with an average of 71 frames for the iLIDS-VID dataset.

Since there is no standard data split, when evaluating on a given dataset, the data is randomly split into training and test set so that each set contains a distinct half of the identities, and the training and evaluation is repeated for 20 trials. This is the usual testing protocol for these datasets \cite{wang2014person}, although the number of trials that is typically used is 10. Here we perform 20 trials for extra stability in our results.

\textbf{Raw input data.} The data used for all experiments is the color and optical flow data (horizontal and vertical), so in practice each frame in a video is represented as a 5-channel image. The optical flow was extracted as a pre-processing step with the Lucas-Kanade algorithm \cite{lucas1981iterative} and normalized to the range $[-1,1]$ as in \cite{McLaughlin16}.

\textbf{Evaluation metric.}
As is common practice for these datasets, the reported metric is the Cumulated Matching Characteristics (CMC) curve, averaged over the chosen number of trials.

\textbf{Dropout.} The networks are trained using dropout \cite{srivastava2014dropout} for regularization purposes. Fig. \ref{fig_seqprocarchitecture} does not show how it is handled during training. In the RNN architecture introduced by \cite{McLaughlin16}, two dropout layers (with same dropout probability of 0.6) were inserted before the input of the first and second fully connected layers $W_i$ and $W_s$ pictured in Fig. \ref{fig_seqprocarchitecture}. In order to keep the FNN architecture as similar as possible, dropout layers with the same probability are inserted at these locations as well.

\subsection{Influence of the recurrent connection}

In this section, we investigate the validity of the simplifying assumptions made in Section \ref{sec_simplification}. In other words, we want to see whether the approximated descriptors defined in equation \eqref{eq_meanpoolapprox} yield a similar performance as the original descriptors for the re-identification task.

For this experiment, we first train a network with our baseline architecture (where the sequence processing stage is a RNN) and we substitute the RNN to use a FNN architecture instead, without changing the values of the weights. Note that this is possible because there is a natural one-to-one mapping between the parameters of the two architectures (see Fig. \ref{fig_seqprocarchitecture}).

\begin{figure}
\includegraphics[width=\linewidth]{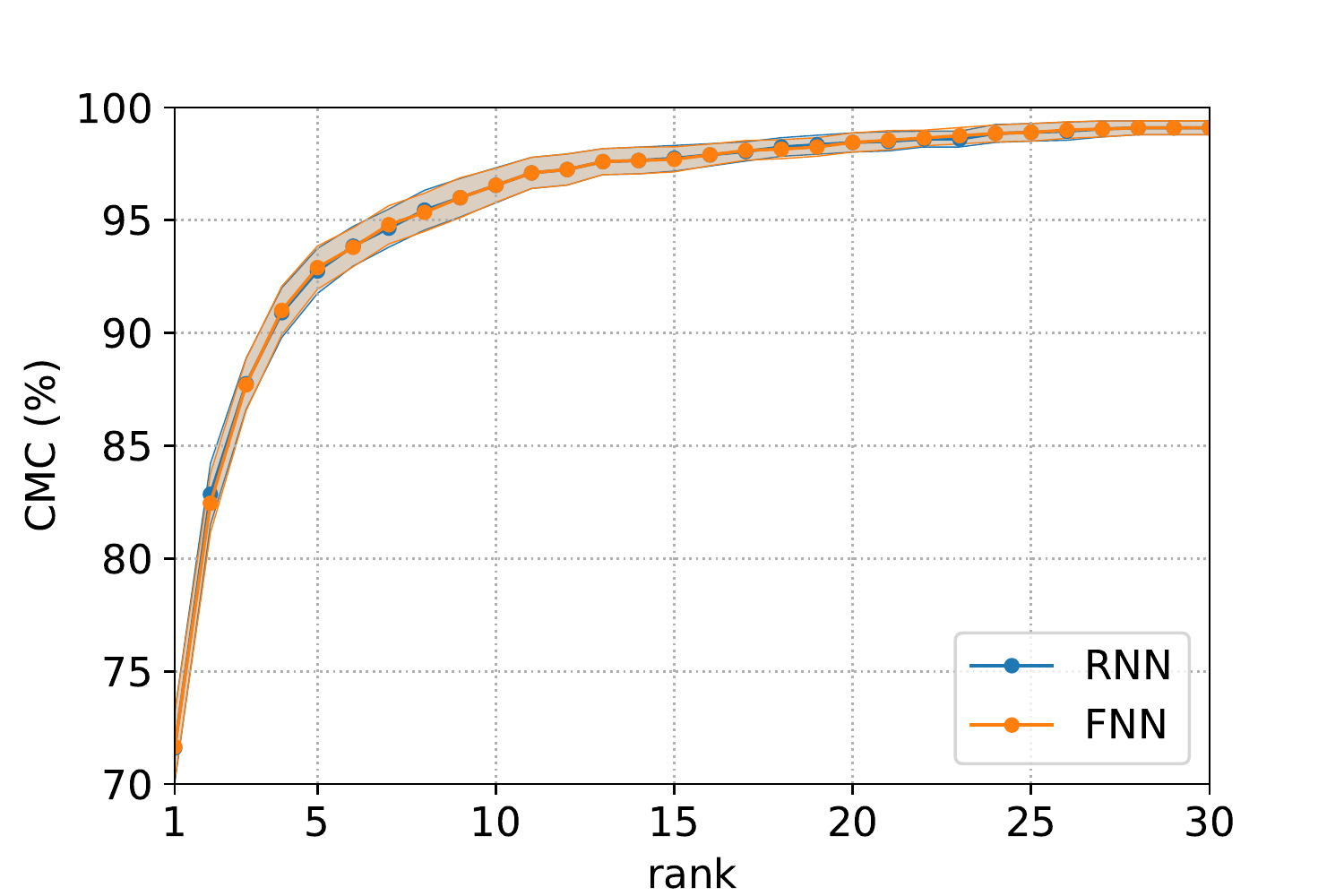}
\includegraphics[width=\linewidth]{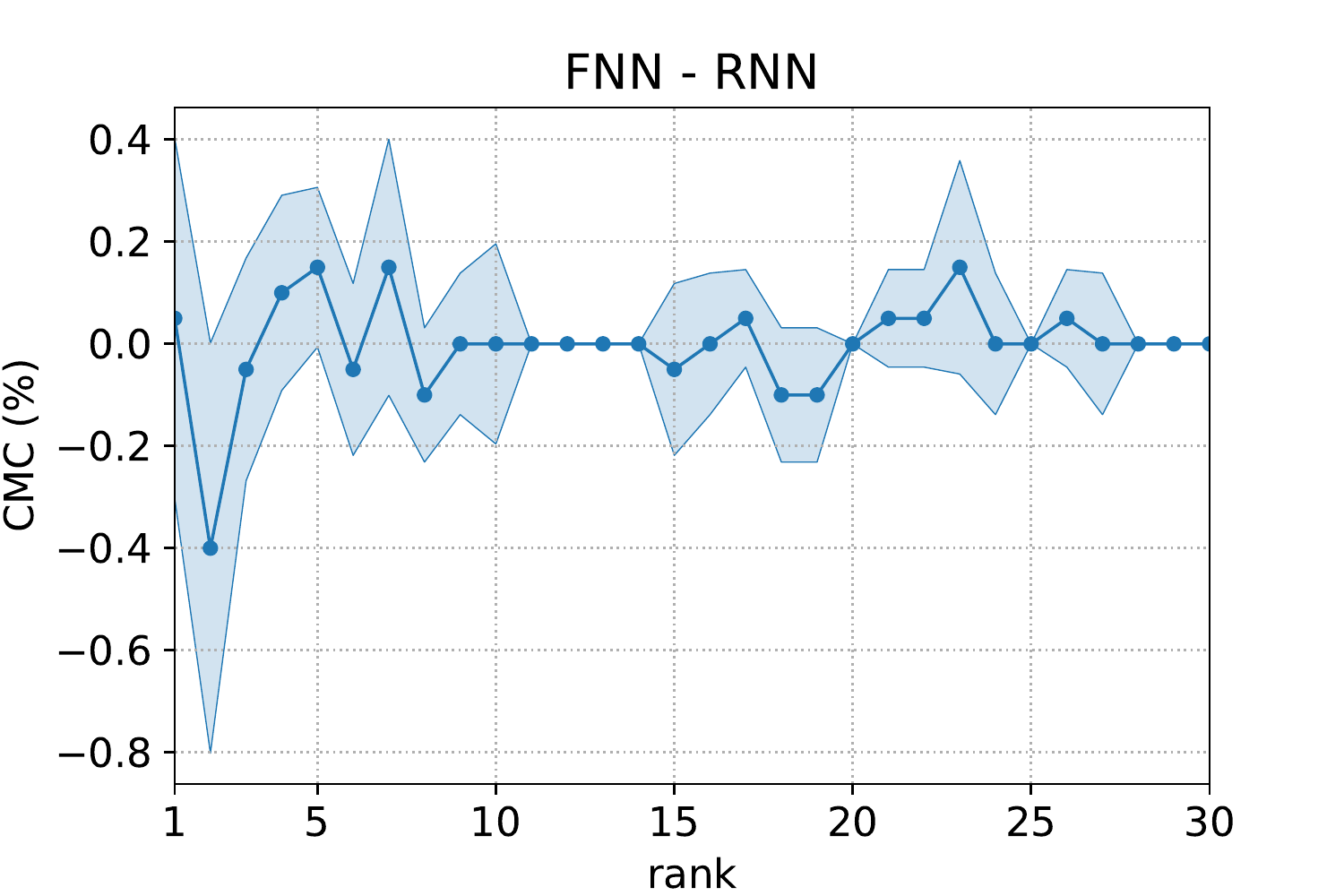}
%\vspace{-6mm}
\caption{Performance of networks with two different architectures for the sequence processing stage (RNN and FNN) but using the same weights. The weights are trained using the RNN architecture. Results are presented on the PRID2011 dataset, randomly split into a training and test set, over 20 trials. 95\% confidence intervals are shown.}
\label{fig_archsubstitution}
\end{figure}

The performance of the two architectures can be compared by evaluating them on the same test set, and comparing their CMC curves. We show the results on the PRID2011 dataset across 20 trials in Fig. \ref{fig_archsubstitution}.
%The plot shows the average CMC value (along with 95\% confidence intervals) for both RNN and FNN architectures.
The first plot shows the average CMC value (along with 95\% confidence intervals) for both RNN and FNN architectures while the second plot shows the distribution of the difference between the CMC values for both architectures.

The average CMC curves for both architectures are very similar, and the difference in the CMC values at each trial is usually very close to 0, with 0 being included in the 95\% confidence intervals for all rank values. If the same weights are used, there is no value of the rank where one architecture consistently outperforms the other one in a statistically significant manner. This means that, out of the box, \emph{removing the recurrent connection does not have any measurable effect on the performance}.

This is a critical result because it challenges the assumptions that are typically made regarding the reasons why a recurrent architecture gives improved performance. It is commonly assumed that the structure of the RNN allows for non-trivial temporal processing of an ordered input sequence, but what we show here is that the performance can be replicated with a non-recurrent network that processes inputs individually and just aggregates them naively, without taking into account the temporal relationship between frames. Surprisingly, the ordering of the sequence of frame features does not matter.

\subsection{Comparison of training modes and architectures}

In this section we provide quantitative evidence that, with the right modifications to the training process, training the feature extraction network without the recurrent connection in the sequence processing stage can boost the re-identification performance, in terms of accuracy as well as speed.

\subsubsection{Training implementation} \label{sec_trainevalsettings}

Our proposed architecture (FNN) is trained using either of the two different training modes as described in section \ref{sec_trainingmodes}. The hyperparameters have to be carefully chosen to keep a fair comparison between the modes.

\textbf{SEQ.} For the first mode, called SEQ, we use the values of hyperparameters reported in \cite{McLaughlin16}, themselves based on \cite{mclaughlin2015data}. The feature dimension is set to 128 and the margin in the contrastive loss function to 2. The inputs are extracted from the full sequences by selecting random subsequences of $L = 16$ consecutive frames. At training time, the inputs are loaded in a mini-batch of size $B = 1$. We alternate between positive and negative pairs of sequences. The network is trained using stochastic gradient descent with a learning rate of $1\mathrm{e}{-3}$. In order to replicate the results reported in \cite{McLaughlin16}, it is necessary to train the network for $1000$ epochs. In this case, an epoch is defined as the number of iterations it takes to show $N$ positive examples ($N$ being the number of identities, or pairs of sequences, in the training set). This ensures that all identities were shown as positive pairs exactly once per epoch. The negative pairs are chosen randomly.

\textbf{FRM.} The FRM training mode, in contrast to SEQ, relies on single frames as an input. This mode uses the same value as SEQ for the feature dimension and margin for the contrastive loss. FRM enables training using much lower memory and computational requirements since the processed input is a pair of images instead of a pair of video sequences. Therefore larger mini-batches can be used. A mini-batch of size $BL$ requires the same resources as a mini-batch of size $B$ for SEQ. So in practice, the comparison between the modes is fair if each FRM mini-batch contains $BL = 16$ pairs of images. These pairs are split in half between positive and negative pairs ($8$ of each). The positive pairs are extracted in a similar fashion as in SEQ: an identity that was not shown yet in the epoch is randomly chosen, and one frame from each of the two camera sequences for that identity is selected. Idem for the negative pairs: two random distinct identities are first chosen and one frame is selected for each of them. It is important to note that an epoch takes fewer iterations to complete for this training mode. In SEQ, the required number of iterations required per epoch is $2 N / B$ (in one epoch $2N$ pairs need to be shown at a rate of $B$ per mini-batch) but in FRM, this number is only $2 N / B L$. So in order to show the same amount of data (same number of mini-batches / iterations for SEQ and FRM) it is necessary to train for $L$ times more epochs in FRM mode. In practice, this means that this network is trained for $16000$ epochs.

It is important to adjust the learning rate in FRM mode. A rule-of-thumb is that when the batch size increases by a factor $k$, the learning rate should increase either by a factor $\sqrt{k}$ \cite{krizhevsky2014one} or more commonly $k$ \cite{krizhevsky2014one, goyal2017accurate}. When switching from SEQ to FRM, the batch size is increased by a factor $L = 16$, so it makes sense to increase the learning rate to some value in the range $[4\mathrm{e}{-3}, 16\mathrm{e}{-3}]$. We found that a good value to use was $16\mathrm{e}{-3}$ for the PRID2011 dataset and $8\mathrm{e}{-3}$ for the iLIDS-VID dataset.

For both training conditions, the input data is diversified by performing data augmentation: random cropping and horizontal mirroring are applied to the inputs. The same augmentation is used for all frames of a subsequence in SEQ.

\subsubsection{Training with SEQ mode}

\begin{figure}
\includegraphics[width=\linewidth]{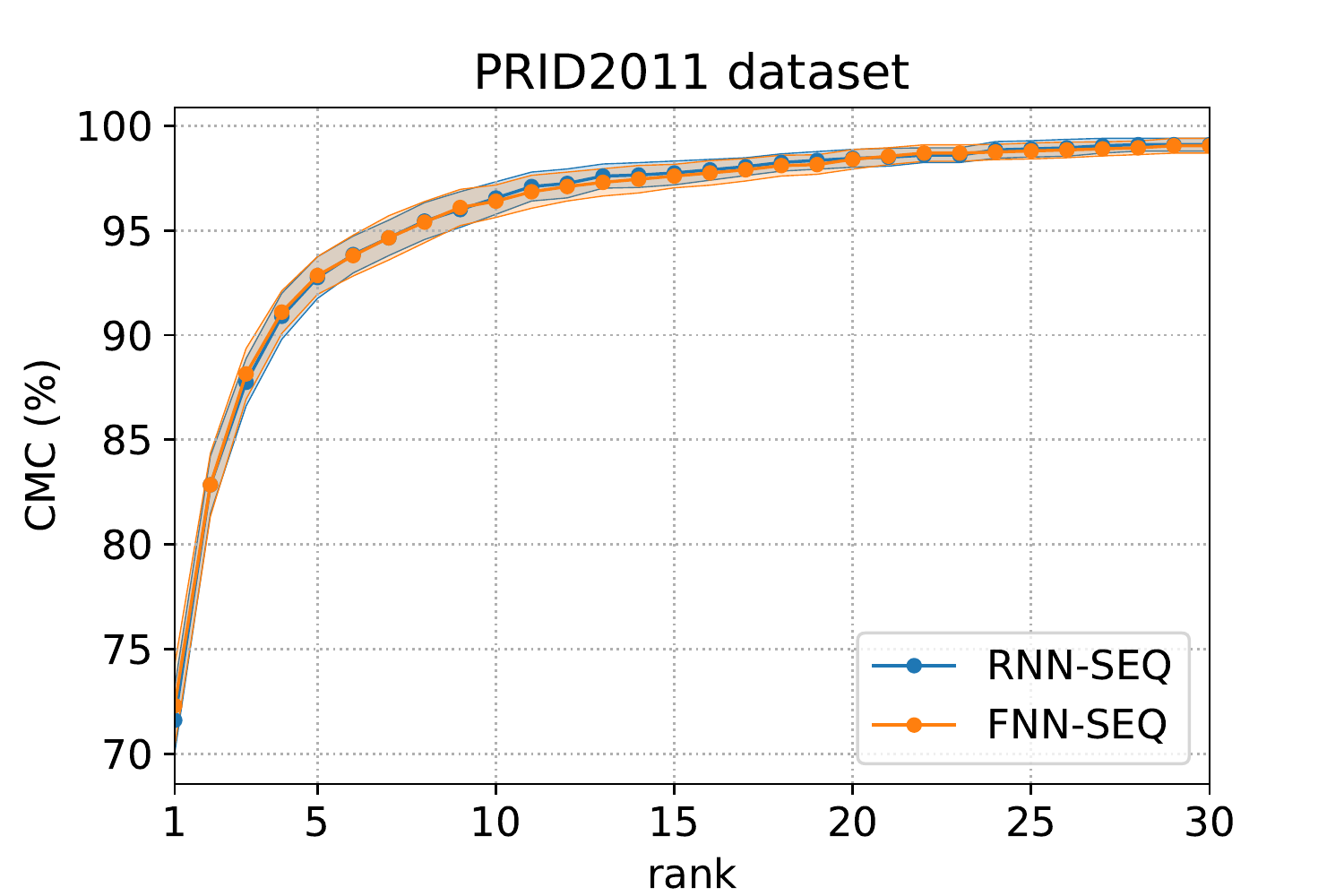}
\includegraphics[width=\linewidth]{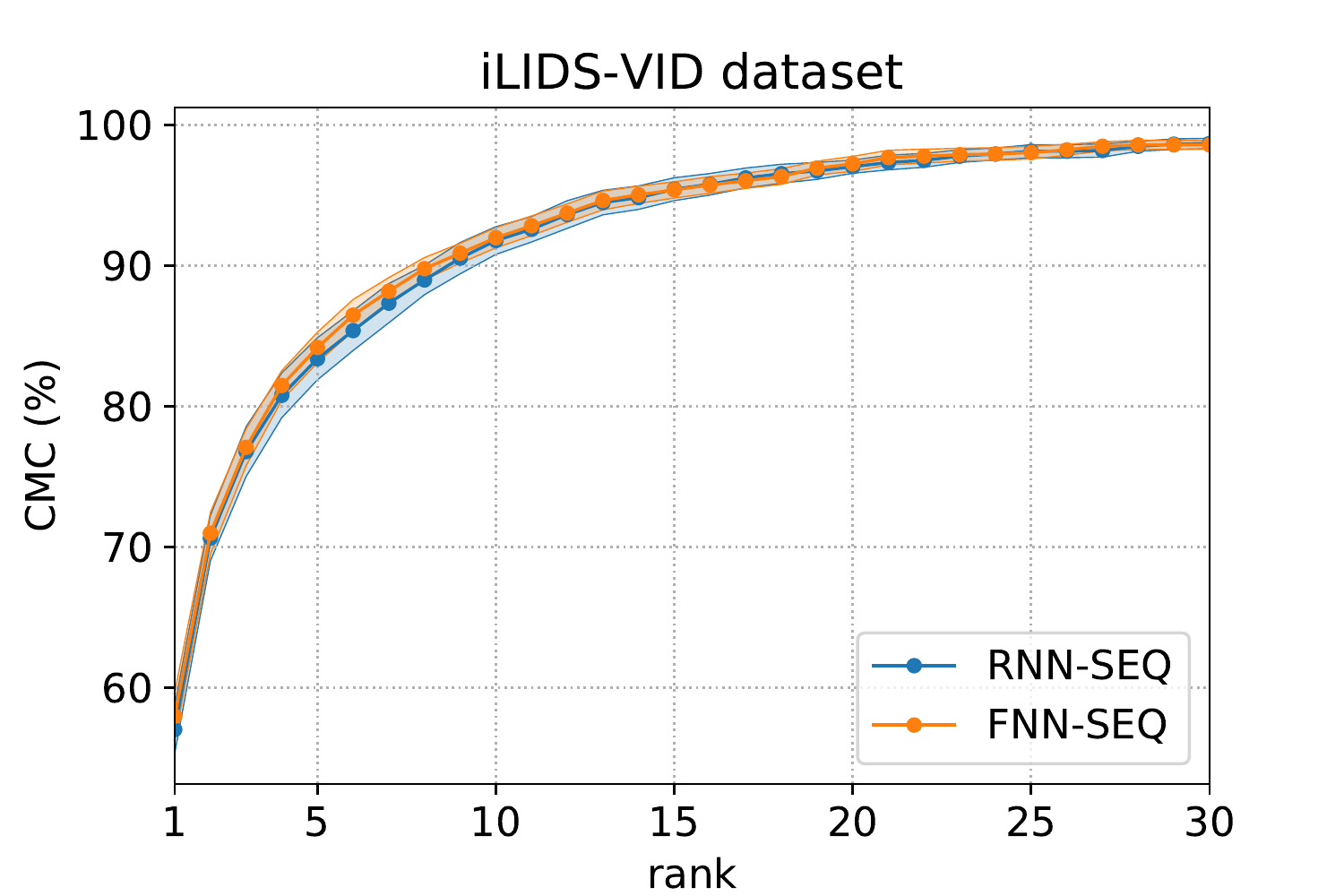}
%\vspace{-6mm}
\caption{CMC curves obtained on each dataset when training a RNN network and a FNN network in SEQ mode. Results are obtained by averaging 20 CMC curves each obtained on a different train/test split. 95\% confidence intervals are shown.}
\label{fig_rnnfnn}
\end{figure}

In this section we compare how the RNN and FNN architectures perform when trained using the SEQ training mode. For each condition we train 20 networks using a different train/test split of the dataset each time.

We report the averaged values as CMC curves (as well as the 95\% confidence intervals) in Fig. \ref{fig_rnnfnn}. As we can see, the values are very similar for both datasets, and the confidence intervals overlap almost perfectly. This shows that the RNN and FNN architectures are functionally equivalent. The good performance obtained by the feature extraction network does not depend on the presence of the recurrent connections in the sequence processing stage, and a feed-forward network that processes frames of a sequence independently, and then uses average pooling across the sequence works just as well.

In \cite{McLaughlin16}, the authors experimented with the removal of recurrent connections in their RNN architecture, and showed a decrease of the retrieval performance as a result. Their conclusion was that these recurrent connections gave the performance boost they observed. What we offer here is a different conclusion. Although it may seem that our results are contradictory, it is actually not the case. In their case, the sequence processing stage was replaced by a single fully connected layer. In our case, the FNN sub-network has a more complex architecture that combines a linear forward path with a nonlinear refinement (Fig. \ref{fig_seqprocarchitecture}). 

\subsubsection{Training with more diverse mini-batches}

\begin{figure}
\includegraphics[width=\linewidth]{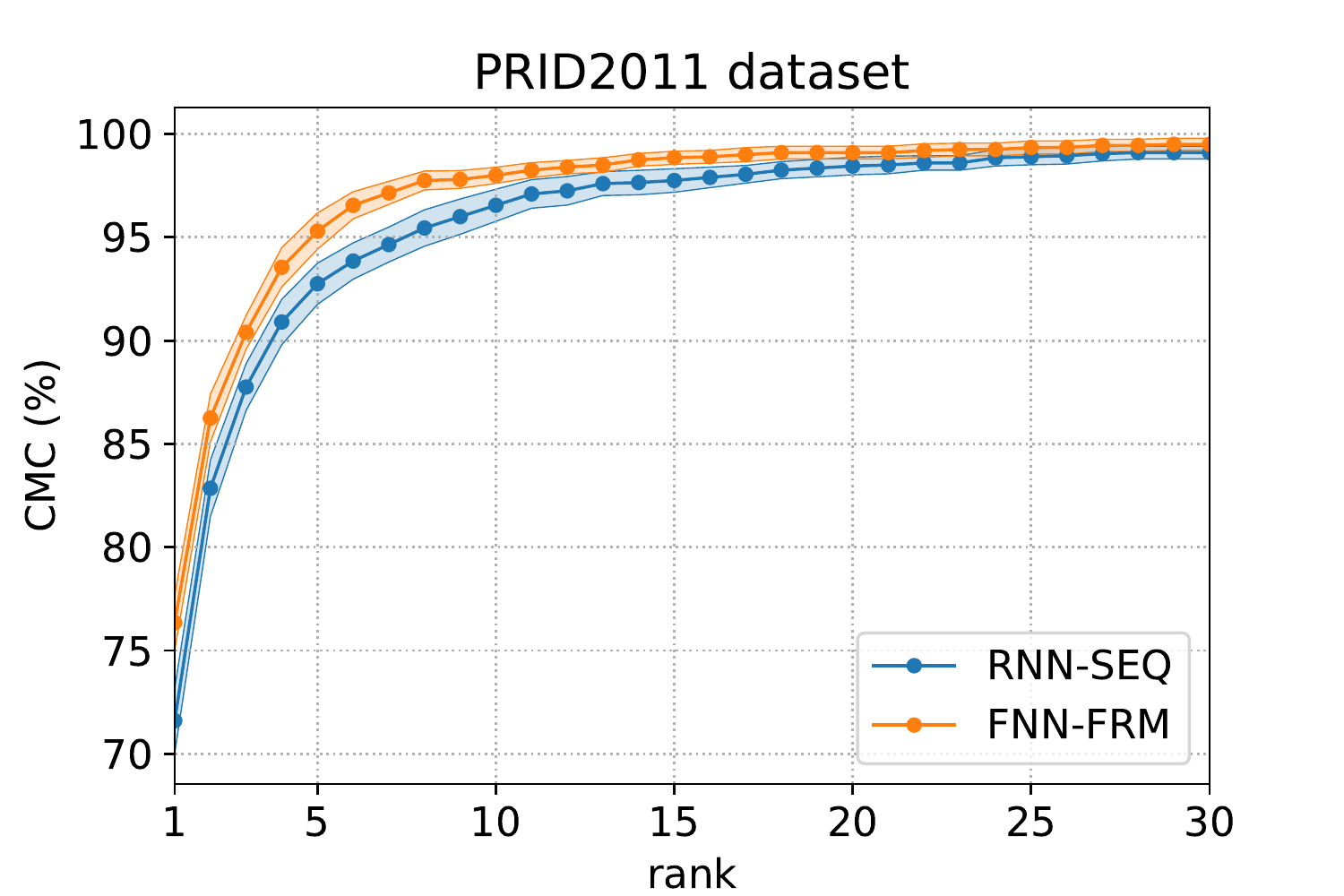}
\includegraphics[width=\linewidth]{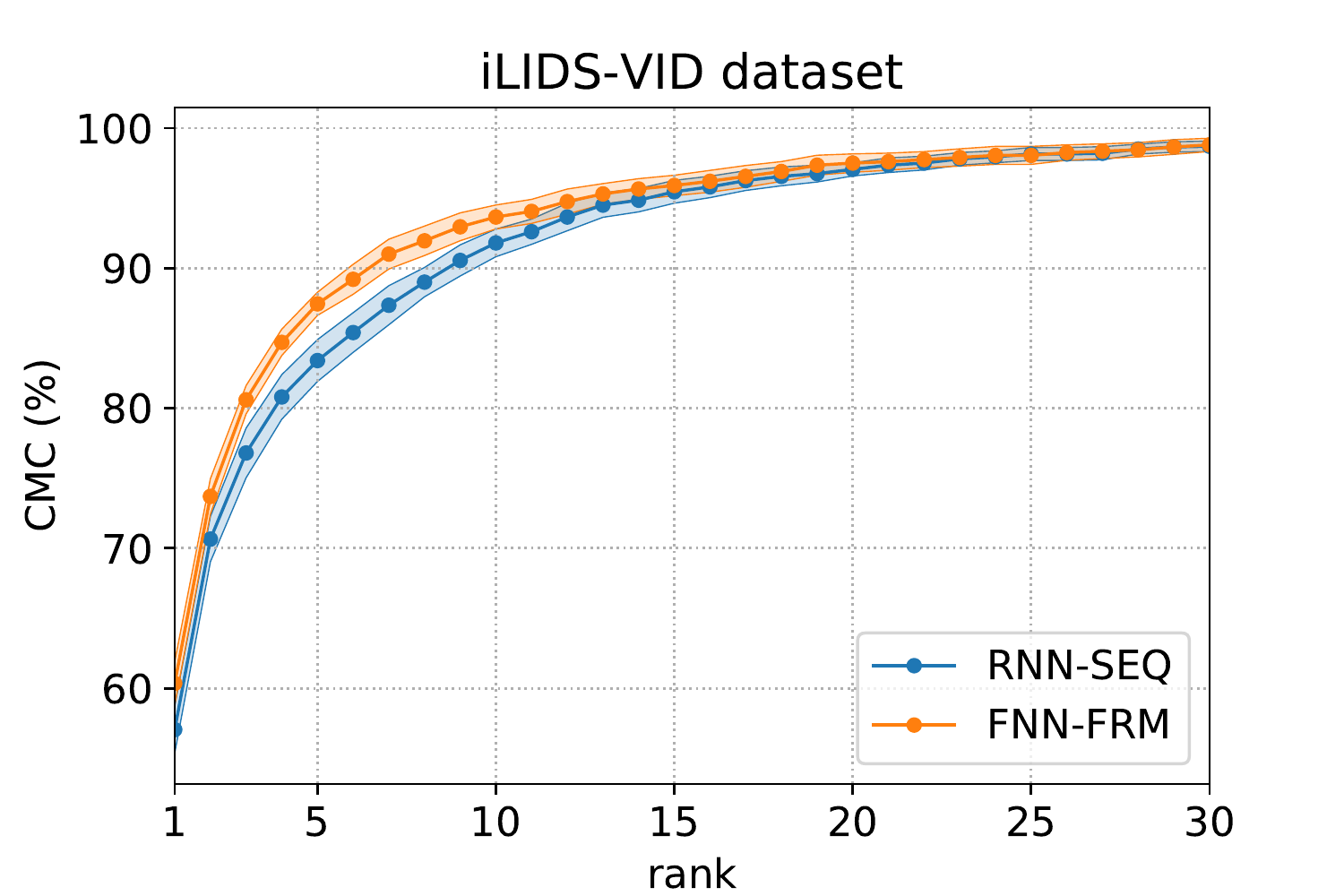}
%\vspace{-6mm}
\caption{CMC curves obtained on each dataset when training a RNN network in SEQ mode and a FNN network in FRM mode. Results are obtained by averaging 20 CMC curves each obtained on a different train/test split. 95\% confidence intervals are shown.}
\label{fig_frmseq}
\end{figure}

\begin{table}
\begin{adjustbox}{center}
\small
\begin{tabular}{lcccccccc}
\toprule
Dataset & \multicolumn{4}{c}{PRID2011} & \multicolumn{4}{c}{iLIDS-VID}\\
\cmidrule(lr){1-1}
\cmidrule(lr){2-5}
\cmidrule(lr){6-9}
Rank & 1 & 5 & 10 & 20 & 1 & 5 & 10 & 20\\
\cmidrule(lr){1-1}
\cmidrule(lr){2-5}
\cmidrule(lr){6-9}
RNN \cite{McLaughlin16}&					70&		90&		95&		97&		58&		84&		91&		96\\
-- (reproduced)&	71.6&	92.8	&	96.6	&	98.5	&	57.1	&	83.4	&	91.8	&	97.1\\
RFA-Net \cite{yan2016person}&	58.2&	85.8&	93.4&	97.9&	49.3&	76.8&	85.3&	90.0\\
Deep RCN \cite{wu2016deep}&		69.0&	88.4&	93.2&	96.4&	46.1&	76.8&	89.7&	95.6\\
Zhou \emph{et al.} \cite{zhou2017see}&		\textbf{\textcolor{blue}{79.4}}&	94.4&	-&		\textbf{\textcolor{blue}{99.3}}&	55.2&	\textbf{\textcolor{cyan}{86.5}}&	-&		97.0\\
BRNN \cite{zhang2017learning}&72.8&	92.0&	95.1&	97.6&	55.3	&	85.0&	91.7	&	95.1\\
ASTPN \cite{xu2017jointly}&	\textbf{\textcolor{cyan}{77}}&		\textbf{\textcolor{cyan}{95}}&		\textbf{\textcolor{blue}{99}}&		99&		\textbf{\textcolor{blue}{62}}&		86&		\textbf{\textcolor{blue}{94}}&		\textbf{\textcolor{blue}{98}}\\
Chen \emph{et al.} \cite{chen2017deep}&		\textbf{\textcolor{cyan}{77}}&		93&		95&		98&		\textbf{\textcolor{cyan}{61}}&		85&		\textbf{\textcolor{blue}{94}}&		97\\
%FNN-SEQ (ours)&			72.3&	92.9&	96.4&	98.4	&	58.0&	84.2&	92.0&	97.3\\
FNN-FRM (ours)&			76.4&	\textbf{\textcolor{blue}{95.3}}	&	\textbf{\textcolor{cyan}{98.0}}	&	\textbf{\textcolor{cyan}{99.1}}	&	58.0	&	\textbf{\textcolor{blue}{87.5}}	&	\textbf{\textcolor{cyan}{93.7}}	&	\textbf{\textcolor{cyan}{97.5}}\\
\bottomrule
\end{tabular}
\end{adjustbox}
\caption{Comparison of our proposed technique against RNN-based methods. Reported results are the mean CMC values (in \%) obtained on the PRID2011 and iLIDS-VID datasets. Best results are shown in dark blue, second best in light blue. Our feed-forward method systematically outperforms its recurrent version \cite{McLaughlin16} and performs better or on par with other RNN-based person re-identification techniques.}
\label{tab_frmseq}
\end{table}

We showed that the performance of the network is the same whether the sub-network in the sequence processing stage is recurrent (RNN) or not (FNN). Without the constraints imposed by the RNN, it makes sense to consider a different training process. Here we examine how the performance of the FNN varies if the training is performed in FRM mode, compared to our baseline (RNN trained in SEQ mode). Again, the results for 20 trials are shown in Fig. \ref{fig_frmseq}. Values for rank 1, 5, 10 and 20 are also given in Table \ref{tab_frmseq}.

For the PRID2011 dataset, the results are very clear: training using single frames instead of sequences gives a noticeable boost. Indeed, for all values of the rank $k$, the average CMC values always exceed (or equate) the baseline values, and the 95\% confidence intervals have no overlap for $k \leq 19$ (except for a very small overlap for $k = 13$).

The results for the iLIDS-VID dataset are also encouraging. There is an improvement of the mean performance for all values of the rank $k \leq 24$, and the 95\% confidence intervals have no overlap for $k \leq 10$. For the very small minority values of $k$ where the baseline performs slightly better than FNN-FRM, the difference never exceeds a statistically insignificant 0.1\%.

The conclusion of this experiment is that switching from SEQ to FRM improves the quality of the training, and in fact it even gives a significant additional boost on both datasets for some values of $k$. This shows the importance of having more diverse data within a mini-batch.

This raises another limitation of using a sequence to train the network. A subsequence of $L$ consecutive frames from a video of a pedestrian will have a large amount of redundancy so the frame features $f^{(t)}$ will have a high degree of similarity between each other. In our FNN formulation, the output of the sequence processing stage for each time step $o^{(t)}$ only depends on $f^{(t)}$, so these outputs will also be very similar. So it is expected that after average pooling, the feature representation of a sequence (mean of all the $o^{(t)}$, $t = 1,...,T$) will not be substantially different from the feature representation of a single frame (any $o^{(t)}$). In the extreme case, if a person does not move at all and the input images are all the same, then all $o^{(t)}$ will be identical and the descriptor for the sequence will also be equal to all $o^{(t)}$. In general, a single frame may yield a descriptor that is slightly more noisy than the descriptor for a sequence, but it should still be sufficient to train our network, while reducing a lot of the wasteful computational cost when computing features for many frames of the same sequence. In fact, we could consider the noise added by computing a descriptor from a single frame compared to a subsequence as some form of data augmentation.

\subsubsection{Speeding up training}

\begin{figure}
\includegraphics[width=\linewidth]{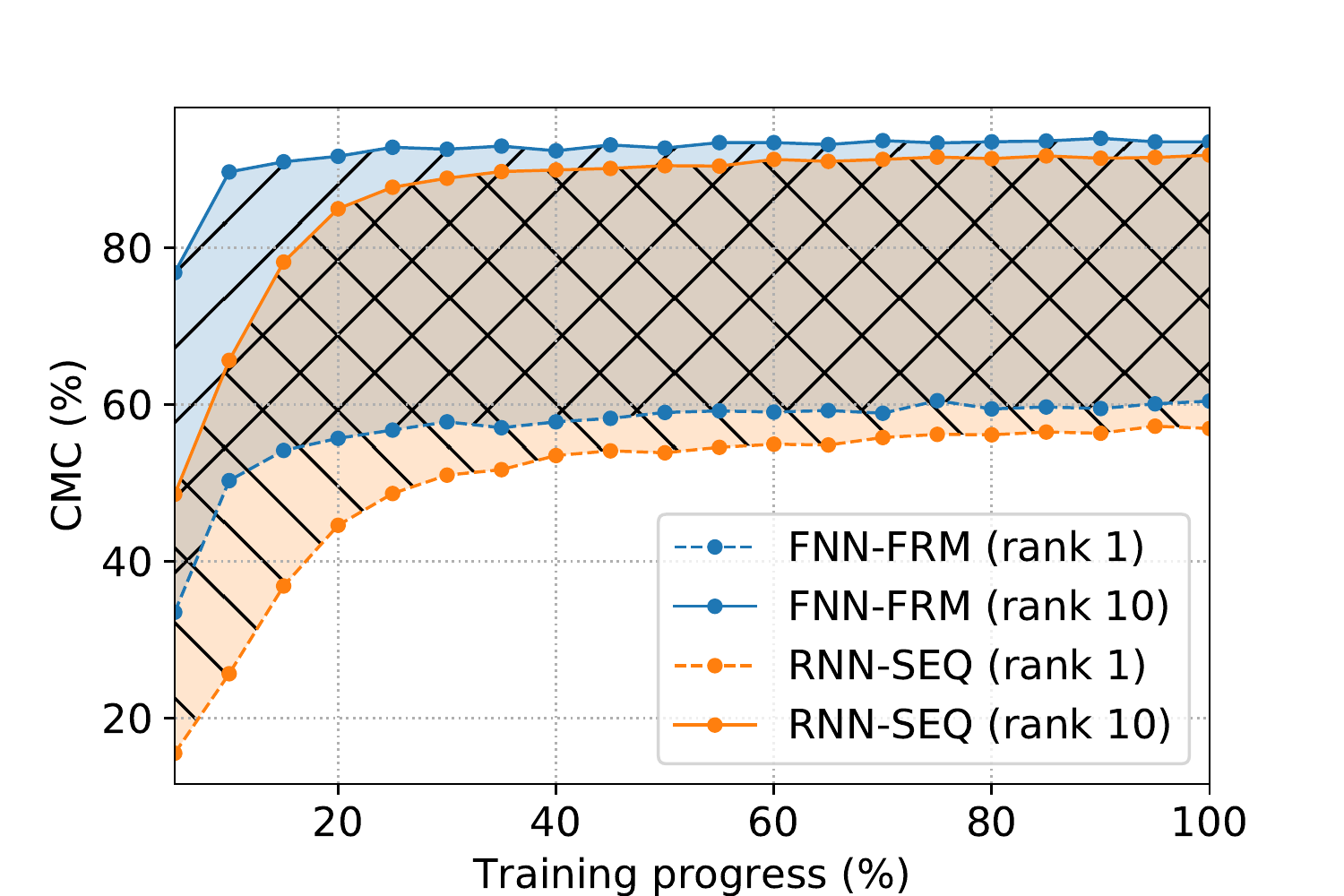}
%\vspace{-6mm}
\caption{Evolution over the training phase of the average CMC values on the test set across training for 20 trials on the iLIDS-VID dataset at rank 1 and rank 10, for both RNN-SEQ and RNN-FRM.}
\label{fig_testhistory}
\end{figure}

One last advantage of FRM over SEQ is that as mentioned in \ref{sec_trainevalsettings} each epoch takes fewer iterations for FRM than for SEQ ($L$ times fewer), so the training accuracy improves much faster. See Fig. \ref{fig_testhistory} for the evolution over time of the CMC values for rank 1 and rank 10. Note that the training progress shown on the x-axis is proportional to the number of iterations, which by design is the same for FRM and SEQ. We can see that after around 25\% of the iterations, FRM already converged and the performance does not noticeably improve after that, while SEQ requires the full number of iterations to fully converge.

To summarize, not only does FRM converge to a higher performance than SEQ, it also gets there much faster.

\subsubsection{Comparison with other RNN-based methods}

Table \ref{tab_frmseq} compares our proposed approach with recent RNN-based person re-identification techniques. We note that FNN trained with FRM always outperforms the baseline (RNN, trained with SEQ). These results also show that our architecture performs better or on par with other methods, that are usually considerably more complex.

\section{Conclusion and future work}

In this work we revisited the use of RNNs in the context of video person re-identification. After deriving a non-recurrent approximation of a simple RNN architecture, we demonstrated experimentally that training this proposed feed-forward architecture could reach very similar performance. Furthermore, we also showed that by training our model on individual frames instead of sequences of frames, we could significantly improve the performance. This also allows for substantially faster convergence than for the recurrent model used as a baseline. The performance we achieve is better or on par with other RNN-based person re-identification techniques.

In future work we plan to explore and experiment on other video-level recognition problems, such as action recognition or video-based object recognition. Indeed, these tasks can also be framed as a high-dimensional embedding of a time-dependent image sequence, which is conceptually not different from what is performed in this paper. Some works in action recognition showed some improvements from previous state-of-the-art methods by using recurrent neural networks, and it would be interesting to explore whether simple FNNs can replace RNNs for these applications as well.

\noindent\textbf{Acknowledgements.}
We thank the authors of \cite{McLaughlin16} for making their code public and for helping us replicating their results.